\documentclass[sigconf,nonacm,balance=false]{acmart}
\usepackage{tabularx,booktabs,array,fvextra,placeins,colortbl}
\usepackage{algorithm,algpseudocode}
\makeatletter
\patchcmd{\maketitle}{\@specialsection{\keywordsname}}
  {\par\smallskip\noindent\textbf{\keywordsname:}\enspace}
  {}{\ClassError{memsidecar}{Cannot format keywords}{Check the acmart keyword block.}}
\makeatother
\definecolor{MSBest}{HTML}{922F27}
\definecolor{MSBestFill}{HTML}{F8EAE7}
\definecolor{MSSecond}{HTML}{245D8A}
\definecolor{MSSecondFill}{HTML}{EAF1F8}
\definecolor{MSGroupFill}{HTML}{EDF1F5}
\definecolor{MSHeader}{HTML}{000000}
\definecolor{MSDeltaGood}{HTML}{187344}
\definecolor{MSDeltaBad}{HTML}{B3261E}
\newcommand{\MSnum}[2]{%
  \begingroup\setlength{\fboxsep}{1.2pt}%
  \ifcase#1\relax
    \hspace{\fboxsep}\makebox[2.75em][c]{#2}\hspace{\fboxsep}%
  \or
    \colorbox{MSBestFill}{\makebox[2.75em][c]{\textcolor{MSBest}{\textbf{#2}}}}%
  \or
    \colorbox{MSSecondFill}{\makebox[2.75em][c]{\textcolor{MSSecond}{#2}}}%
  \fi\endgroup}
\floatstyle{ruled}
\restylefloat{algorithm}
\providecommand{\tightlist}{\setlength{\itemsep}{0pt}\setlength{\parskip}{0pt}}
\setcopyright{none}
\title{Who Schedules Agent Memory? A Sidecar Approach to Long-Horizon Tasks}
\author{Qijia Zhuang}
\affiliation{%
  \institution{Nanjing University}
  \city{Nanjing}
  \country{China}}
\author{Rui Liu}
\affiliation{%
  \institution{Northeastern University}
  \country{China}}
\author{Yuxiang Ren}
\authornote{Corresponding author: \href{mailto:renyuxiang@nju.edu.cn}{renyuxiang@nju.edu.cn}.}
\affiliation{%
  \institution{Nanjing University}
  \city{Nanjing}
  \country{China}}
\email{renyuxiang@nju.edu.cn}
\hypersetup{keeppdfinfo=true,pdfauthor={Qijia Zhuang, Rui Liu, and Yuxiang Ren}}
\makeatletter
\def\@mkauthors{%
  \gdef\@currentauthors{}%
  \gdef\@currentaffiliation{}%
  \global\setbox\mktitle@bx=\vbox{%
    \unvbox\mktitle@bx\par\medskip
    \centering
    {\@authorfont Qijia Zhuang\textsuperscript{1},
      Rui Liu\textsuperscript{2},
      Yuxiang Ren\textsuperscript{1,*}\par}%
    \smallskip
    {\@affiliationfont
      \textsuperscript{1}Nanjing University, Nanjing, China\par
      \textsuperscript{2}Northeastern University, China\par}%
    \bigskip}}
\makeatother
\begin{document}
\begin{abstract}
As large language model agents increasingly undertake complex, long-horizon tasks, memory has become an important component for supporting sustained decision-making and action. Existing methods primarily focus on memory storage, updating, and retrieval, but often overlook how memory scheduling affects task execution: the main agent must identify memory needs, initiate retrieval, and wait for results while advancing the task, increasing both scheduling burden and execution time. To address this problem, we propose MemSidecar, which decouples memory scheduling from task execution through two independently selectable modes. Hooks directly delivers bound memory to the main agent when execution events match predefined conditions. Foresight uses an independent agent to predict potential queries, retrieve results in parallel, and store them in a cache, delivering matching and valid results when actual requests arrive. Hooks and Foresight achieve scores of 41.23\% and 44.34\% on MCP-Persona, respectively, while effectively reducing execution costs and latency and improving the efficiency of long-horizon agent execution.
\end{abstract}
\maketitle
\section{Introduction}\label{introduction}

Large language model agents are moving from answering questions to sustained task execution: browsing requires combining clues across pages, application use requires connecting successive operations, and project work requires carrying earlier decisions and progress forward \cite{yao2023react,jimenez2024swebench}. As task horizons grow, continuity across steps and sessions becomes an important basis for sustained execution. For Web application tasks, we study memory scheduling during long-horizon agent execution so that relevant memories can be delivered automatically and continually support subsequent decisions and actions.

Memory provides historical information that remains available throughout long-horizon tasks. Prior studies retain histories, summarize experience, and recover information across sessions to support continued decisions \cite{park2023generative,maharana2024locomo}. Once information is stored, it must also be retrieved and delivered at the appropriate execution point.

This raises two related problems.

\textbf{First, retrieval timing and waiting overhead.} In a typical on-demand workflow, the main agent identifies memory needs from the current instruction or query, then constructs a retrieval query and accesses memory \cite{packer2023memgpt}. In long-horizon tasks, identifying access needs only from the current input may overlook earlier constraints, task state, or cross-task dependencies, or delay retrieval until after the relevant decision. Waiting for memory recall blocks task progress and increases overall execution latency; repeated accesses bring demand identification, query construction, and retrieval waits into the task's critical path \cite{chhikara2025mem0}.

\textbf{Second, retrieval coverage.} In long-horizon tasks, the current instruction may not explicitly contain all the historical cues needed to complete the task. Relevant constraints, prior progress, and cross-task dependencies may be distributed across memories and expressed differently from the current instruction. Constructing queries solely from that instruction may therefore miss relevant information, making task goals and visible context necessary for determining what to retrieve. As a task unfolds, a planning step may need long-term constraints, whereas a later tool action may depend on recent progress or an earlier artifact. The relevant memory set therefore changes with execution, even when the user does not introduce a new instruction.

Existing memory research emphasizes storage, updating, and retrieval. In a typical on-demand interface, the main agent still initiates memory access inside its task loop. We study this scheduling coupling: the agent makes task decisions while identifying memory needs, constructing queries, and initiating and waiting for memory access. This adds scheduling burden to the main agent, while memory recall waits block task progress and increase overall execution latency. We assign memory scheduling to a separate module that uses task goals and visible context to determine retrieval content and organizes triggering, preparation, and delivery around task execution.

Figure 1 illustrates an XH hotel-booking task resumed after earlier sessions established the budget and meeting arrangements. It shows retrieval not being triggered, blocking from serial memory access, and incomplete queries. Stored information may still fail to reach the agent before the relevant decision.

\begin{figure}[t]
\centering
\includegraphics[width=\columnwidth]{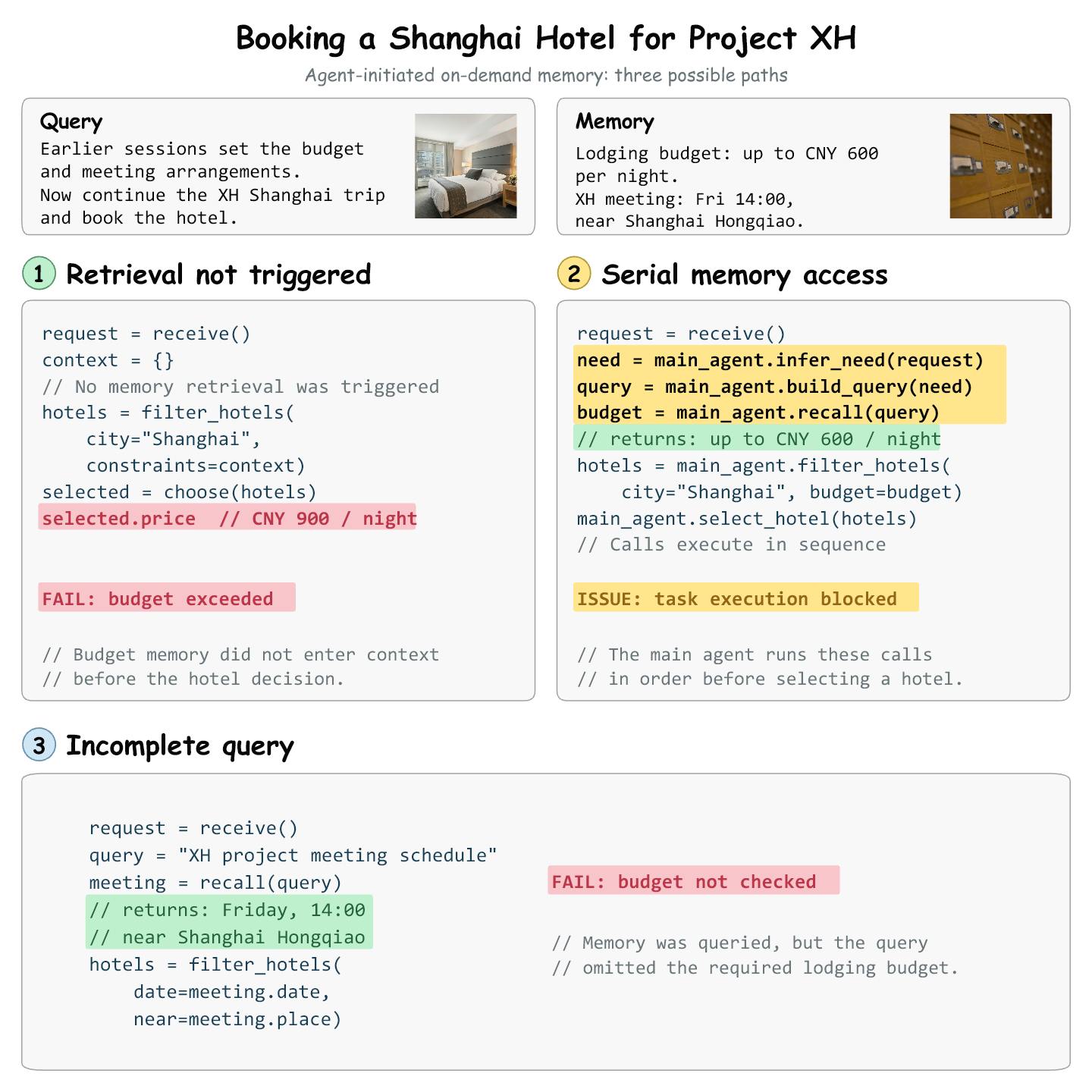}
\caption{Three failure modes in the XH hotel-booking task.}
\label{fig:1}
\end{figure}

We propose \textbf{MemSidecar} with two complementary memories. Global Memory organizes knowledge and user preferences reusable across tasks. Workspace Memory organizes knowledge, skills, experience, and task state within a particular workspace. In a multi-step workflow, a standing budget can remain applicable while selected resources, completed operations, and unresolved dependencies evolve in the workspace. Recovering both kinds of information gives a resumed task an explicit account of its requirements and progress. The scheduling module can then associate a decision with the relevant constraint or prior result as that decision approaches.

\begingroup
\brokenpenalty=100\relax
MemSidecar offers two independently selectable modes, \textbf{Hooks} and \textbf{Foresight}. Hooks binds specific memory entries to executable condition functions generated by an LLM. During execution, the system evaluates these functions against the current event, state, or pending action. When a condition holds, it delivers the corresponding memory directly to the main agent without requiring the agent to initiate a memory query. Foresight uses a memory agent to predict likely queries from the initial instruction and memory topic directory, retrieve results while the main agent executes, and cache them for subsequent requests.
\par
\endgroup

Taken together, our main contributions are:

\begin{itemize}
\tightlist
\item
  We formulate retrieval timing, demand identification, and execution blocking as memory-scheduling problems in long-horizon agents and design two sidecar modes, Hooks and Foresight.
\item
  We provide a unified Global and Workspace Memory interface that organizes cross-task knowledge and user constraints by topic and supports continued execution with workspace-specific skills, experience, and task state.
\item
  Evaluation on MemoryArena and MCP-Persona shows that MemSidecar improves task accuracy while increasing execution efficiency.
\end{itemize}

\section{Related Work}\label{related-work}

\subsection{Agent Memory}\label{agent-memory}

Long-horizon tasks require using historical information across steps, for example by integrating previously gathered clues during persistent browsing. However, models' use of contextual information still depends on information position and task complexity; larger context windows do not ensure effective use of their contents \cite{liu2024lost,hsieh2024ruler}.

Existing research primarily studies how to store, organize, and update memory. Historical events and user characteristics can be retained through continual updates \cite{zhong2024memorybank}, while task feedback can be distilled into experience for subsequent attempts \cite{shinn2023reflexion}. Knowledge graphs connect pieces of information to support retrieval and integration of scattered clues \cite{gutierrez2024hipporag}; hierarchical context and multistage consolidation manage information across timescales \cite{packer2023memgpt,fang2026lightmem}. State memory further preserves execution progress and tool evidence, while learned maintenance improves memory updates for subsequent tasks \cite{li2026flowstate,zhang2026uma}. Procedural memory also distills successful executions into reusable skill compositions and uses task state to guide skill selection \cite{zhou2026spem}.

Memory use also depends on current decisions and access costs. Systems research examines the connections between memory, actions, and decisions, as well as the costs of representation, extraction, routing, and maintenance \cite{sumers2023coala,zhou2026agentnative}. Overuse or underuse of memory and differences in retained history also affect task performance \cite{cao2026memcalib,huang2026retention}. For Web tasks, Synapse stores successful interaction trajectories as exemplar memory and retrieves similar exemplars to guide subsequent actions \cite{zheng2024synapse}; Agent Workflow Memory induces reusable workflows from past interactions and incorporates them into memory to guide subsequent web tasks \cite{wang2025agentworkflow}. Alongside what to retain, it is therefore necessary to consider when this information can support the current task.

MemSidecar uses Global Memory to store knowledge and user preferences across tasks, and Workspace Memory to store workspace experience and task state. We further study when to prepare and deliver these two types of memory: what information current and subsequent steps need, and when to provide it to the main agent. Hooks triggers memory delivery in response to events, while Foresight predicts needs and prepares memory in advance, allowing memory access to support subsequent decisions as tasks progress.

\subsection{Information Retrieval and Scheduling}\label{information-retrieval-and-scheduling}

Retrieval scheduling connects need formation, access timing, and result delivery. Retrieval-augmented generation integrates external information into generation, while black-box augmentation supports configuring a retriever outside a frozen model \cite{lewis2020rag,shi2023replug}; scheduling also changes the distribution of construction, retrieval, and generation costs \cite{omri2026agentmemory}.

Prior work forms queries as tasks progress by decomposing follow-up questions or interleaving reasoning with retrieval \cite{press2023selfask,trivedi2023ircot}. FLARE anticipates information needs from predicted continuations \cite{jiang2023flare}. Self-RAG uses reflection tokens to guide retrieval \cite{asai2023selfrag}. Adaptive-RAG selects retrieval strategies by question complexity \cite{jeong2024adaptiverag}, while CRAG evaluates retrieved evidence and chooses corrective actions \cite{yan2024crag}. Together, these mechanisms place retrieval decisions both before information access and after the initial evidence has returned. Query construction and triggers jointly shape timely access. MemSidecar places memory scheduling alongside execution, using event-triggered Hooks or independently predictive Foresight to study when task memory is prepared and returned to execution.

\section{Method}\label{method}

MemSidecar assigns memory scheduling to a sidecar module working with the main agent. The main agent plans and executes the task; the sidecar organizes history, matches memory triggers, and prepares context. Global Memory and Workspace Memory provide shared storage for two independently selectable scheduling modes, Hooks and Foresight.

\begin{figure*}[t]
\centering
\includegraphics[width=\textwidth]{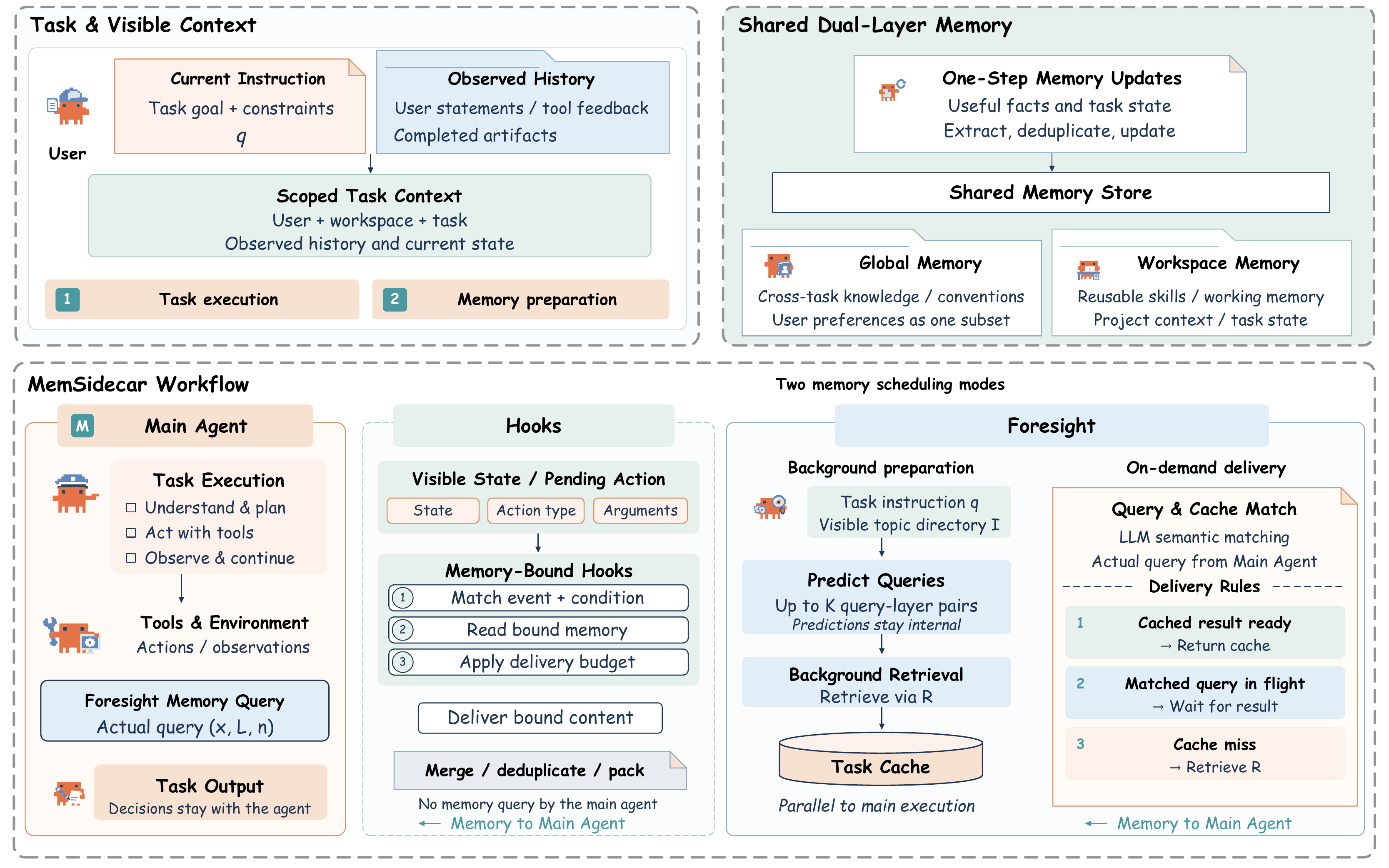}
\caption{MemSidecar framework. Global Memory organizes knowledge and user preferences reusable across tasks; Workspace Memory organizes workspace-specific skills, experience, and work state. Hooks directly delivers bound memory when execution state or a pending action satisfies its condition; Foresight predicts queries from the task instruction and prepares a task-level cache in the background. The main agent uses the delivered context for task decisions.}
\label{fig:2}
\end{figure*}

\subsection{Problem Formulation}\label{problem-formulation}

Given instruction \(q\), the main agent completes a task through successive actions. Its state at step \(t\) is \(s_t=(q,H_t,o_t)\), where \(H_t\) is the within-task history sequence and \(o_t\) the current observation. It generates an action using delivered memory context \(r_t\):

\begingroup
\def\theHequation{memsidecar-1}
\begin{equation}
a_t=\pi_{\mathrm{main}}(s_t,r_t).
\tag{1}
\end{equation}
\endgroup

Actions include environment operations, tool calls, and submitting results. The external store \(\mathcal M\) contains history from before the current task and new information observed during execution.

Long-horizon tasks separate the initial request from the decisions that depend on earlier information. Constraints and prior decisions may become relevant only at later stages, after useful evidence has left the active context. Memory scheduling combines query content with task progress to determine when to recover historical evidence and deliver it to the relevant decision or action.

\textbf{Conventional on-demand scheduling.} The main agent decides whether memory is needed, constructs a query, and continues after retrieval returns. Let \(d_t\in\{0,1\}\) be the access decision and \(x_t\) the query; \(D\) and \(Q\) denote the corresponding decision and query-construction functions. The interface \(R(x,\mathcal L,n;\mathcal M)\) returns entries from the store, with nonempty target layers \(\varnothing\ne\mathcal L\subseteq\{\mathrm{G},\mathrm{W}\}\) (both by default) and result limit \(n\); \(\operatorname{Pack}\) converts entries into context:

\begingroup
\def\theHequation{memsidecar-2}
\begin{equation}
\begin{aligned}
d_t&=D(s_t),\qquad x_t=Q(s_t),\\
r_t&=\operatorname{Pack}\bigl(R(x_t,\mathcal L,n;\mathcal M)\bigr)\quad(d_t=1).
\end{aligned}
\tag{2}
\end{equation}
\endgroup

When \(d_t=0\), \(r_t\) is empty context. Scheduling and query construction consume agent decision steps; synchronous retrieval adds waiting to the critical path. Let \(\mathcal U=\{t:d_t=1\}\) be access steps, \(T_{\mathrm{task}}\) the task's own execution time, and \(T^{\mathrm{sched}}_t\) and \(T^{\mathrm{ret}}_t\) the scheduling and retrieval waits. A serial on-demand workflow is:

\begingroup
\def\theHequation{memsidecar-3}
\begin{equation}
T_{\mathrm{coupled}}=T_{\mathrm{task}}+\sum_{t\in\mathcal U}(T^{\mathrm{sched}}_t+T^{\mathrm{ret}}_t).
\tag{3}
\end{equation}
\endgroup

Repeated retrieval entails query construction, search, and context processing. Repeated delivery consumes input tokens and context space, and recall waits block subsequent actions. These costs accumulate with trajectory length.

\textbf{Sidecar scheduling.} MemSidecar uses an independent scheduler \(S\) to prepare memory from the instruction and visible event history \(e_{\leq t}\) through step \(t\). Events convey relevant information about the current state and pending action:

\begingroup
\def\theHequation{memsidecar-4}
\begin{equation}
r_t=S(q,e_{\leq t},\mathcal M),\qquad a_t=\pi_{\mathrm{main}}(s_t,r_t).
\tag{4}
\end{equation}
\endgroup

Hooks directly delivers bound memory under state or action conditions, whereas Foresight prepares query results through predictive prefetching. Both organize content in the same dual-layer store, and the main agent uses the delivered context for task decisions.

\subsection{Dual-Layer Memory}\label{dual-layer-memory}

The layers distinguish information by its scope of reuse. Global Memory stores knowledge and user preferences reusable across tasks. Workspace Memory stores knowledge, skills, experience, and task state within a particular workspace. They address what knowledge and user preferences are reusable and what experience and progress are available in the current workspace, respectively. The store is \(\mathcal M=(\mathcal M_{\mathrm{G}},\mathcal M_{\mathrm{W}})\). Here \(\mathcal M_{\ell,t}\) is the state of layer \(\mathcal M_\ell\) at step \(t\), with \(\ell\in\{\mathrm{G},\mathrm{W}\}\).

Memory extraction and maintenance. At a stage or task boundary, the sidecar extracts useful facts and task state from observed interactions and tool results. Each entry expresses a reusable fact or a coherent piece of state, together with its applicability conditions. General knowledge and user preferences enter Global Memory; project procedures, completed work, and pending steps enter Workspace Memory.

Extraction, duplicate checking, and necessary updates are performed in one maintenance pass. Existing entries are retained when they already capture the information; new observations add missing facts or revise changed state. For example, a travel entry can record a confirmed booking and its dates, while a project entry records the finished step and next dependency. The trajectory is organized into these fact and state entries, with useful content and applicability conditions preserved.

Global Memory: a topic-organized knowledge graph. Global Memory \(\mathcal M_{\mathrm{G}}\) is represented by graph \(\mathcal G_{\mathrm{G}}\), comprising a topic index and knowledge nodes. Topics provide coarse-to-fine access paths, while knowledge nodes hold content for decisions and its applicability conditions:

\begingroup
\def\theHequation{memsidecar-5}
\begin{equation}
\mathcal G_{\mathrm{G}}=(\mathcal V_{\mathrm{T}}\cup\mathcal V_{\mathrm{K}},\mathcal E_{\mathrm{G}}).
\tag{5}
\end{equation}
\endgroup

\(\mathcal V_{\mathrm{T}}\) and \(\mathcal V_{\mathrm{K}}\) denote the sets of topic and knowledge nodes, respectively; \(\mathcal E_{\mathrm{G}}\) is the relation set between topics and knowledge and among nodes of either kind. Parent--child topic relations form a hierarchical index: a parent topic groups related needs, and its children narrow the access direction. A knowledge node can connect to several related topics, allowing different tasks to reach the same content through their respective topic paths.

Knowledge nodes are the basic units of reuse. Each node organizes content around a specific proposition and records the conditions under which it applies. Knowledge nodes record reusable understanding or user preferences. The topic index determines where to look, while knowledge nodes determine what information to supply.

Global Memory can hold user information, user preferences, and cross-task knowledge, with applicability conditions supporting interpretation and reuse.

Workspace Memory: skills and working memory. The workspace layer combines reusable skills \(\mathcal S\) and working memory \(\mathcal W\):

\begingroup
\def\theHequation{memsidecar-6}
\begin{equation}
\mathcal M_{\mathrm{W}}=(\mathcal S,\mathcal W).
\tag{6}
\end{equation}
\endgroup

\(\mathcal S\) stores established project workflows, experience, and procedures for similar tasks. \(\mathcal W\) stores current task progress and artifacts. The former supports reuse across tasks; the latter supports continuity across steps. Project and topic indices connect them so an agent can recover both how to work and where work has reached.

Memory selection and delivery. Let \(\operatorname{Entries}(\mathcal M)\) be the store's memory-entry set and \(\mathcal B_t\subseteq\operatorname{Entries}(\mathcal M)\) the selected subset. Delivery merges and deduplicates their useful text, then packs it within the context budget:

\begingroup
\def\theHequation{memsidecar-7}
\begin{equation}
r_t=\operatorname{Pack}(\mathcal B_t).
\tag{7}
\end{equation}
\endgroup

For Hooks, \(\mathcal B_t\) contains the bound entries whose conditions match; for Foresight, it contains prefetched or normally retrieved query results. Both use the same content organization and delivery budget.

\subsection{Hooks: Functional Triggers Based on State and Action}\label{hooks-functional-triggers-based-on-state-and-action}

Hooks binds host events and checkable conditions to memory entries. During execution, the sidecar reads relevant state and pending actions; a matching condition directly delivers the bound entry's text at the current decision point. Functional activation draws on HASP's use of state and candidate actions to determine skill applicability \cite{liu2026hasp}. The main agent uses the bound content to continue planning and execution without first issuing a memory query.

Adding a Hook to a memory entry. At memory writing or configuration time, an LLM generates an executable predicate from the entry's content and applicability conditions, then binds it and the corresponding host event to that entry. For example, ``run unit tests before deploying this project'' can bind to a pre-tool event whose condition identifies a deployment action for that project. A match directly supplies the convention so the main agent can check test status first. One entry can carry multiple Hooks for different uses.

Write the event as \(e_t=(\epsilon_t,\sigma_t,\tilde a_t)\), where \(\epsilon_t\) is the event type, \(\sigma_t\) the visible projection of \(s_t\), and \(\tilde a_t\) the proposed action and its arguments, distinct from the final executed action \(a_t\); \(\tilde a_t=\varnothing\) when none is proposed. State conditions check task phases or environmental state; action conditions check the operation and arguments before execution. They can be configured separately or combined.

Let \(\mathcal J=\{1,\ldots,N_H\}\) index the \(N_H\) currently valid Hooks. Each Hook is:

\begingroup
\def\theHequation{memsidecar-8}
\begin{equation}
h_j=(m_j,\mathcal E_j,\chi_j).
\tag{8}
\end{equation}
\endgroup

\(m_j\) is the bound entry, \(\mathcal E_j\) the allowed event-type set, and \(\chi_j(\sigma_t,\tilde a_t)\in\{0,1\}\) its applicability predicate. We use \(h_j(e_t)\) for the trigger mapping induced by this configuration; a match returns the bound entry:

\begingroup
\def\theHequation{memsidecar-9}
\begin{equation}
\begin{aligned}
h_j(e_t)=m_j,\\
 \epsilon_t\in\mathcal E_j,\ \chi_j(\sigma_t,\tilde a_t)=1.
\end{aligned}
\tag{9}
\end{equation}
\endgroup

Matching and context delivery. At the corresponding host events, the scheduler checks Hooks and merges and deduplicates their matched entries:

\begingroup
\def\theHequation{memsidecar-10}
\begin{equation}
\begin{aligned}
r_t&=\operatorname{Pack}\left(\{m_j:j\in\mathcal J_t\}\right),\\
\mathcal J_t&=\{j\in\mathcal J\setminus\mathcal F_t:\epsilon_t\in\mathcal E_j,\ \chi_j(\sigma_t,\tilde a_t)=1\}.
\end{aligned}
\tag{10}
\end{equation}
\endgroup

Here \(\mathcal F_t\) contains the indices of Hooks already successfully delivered in the current task.

State triggers supply memory when the relevant state is observed; action triggers supply it before the corresponding operation executes. The entry binding determines content, and the predicate determines delivery timing. Memory received before execution lets the main agent revise its pending action, connecting the entry to the decision it affects.

\subsection{Foresight: Task-Aware Predictive Prefetch Caching}\label{foresight-task-aware-predictive-prefetch-caching}

Foresight predicts potential queries from the initial instruction and visible memory-topic directory, then prepares corresponding historical results in a task-scoped cache. Background preparation overlaps with the main agent's computation to reduce later access waits \cite{linux2026readahead}.

Predicting accesses. An independent ForesightAgent generates queries from instruction \(q\) and visible directory \(I\). Prediction starts with the task and can supplement preparation after observed memory writes update the directory. Here \(F\) summarizes the cumulative preparation plan, with \(J\) queries across the task and budget \(K\); \(\hat x_i\) is a predicted query and \(\widehat{\mathcal L}_i\) its nonempty target-layer set:

\begingroup
\def\theHequation{memsidecar-11}
\begin{equation}
\widehat{\mathcal Q}=F(q,I,K)=\{(\hat x_i,\widehat{\mathcal L}_i)\}_{i=1}^{J},\qquad J\leq K.
\tag{11}
\end{equation}
\endgroup

The directory provides an index view containing topics, categories, tags, and memory layers. The predictor combines it with the initial instruction to identify potential memory needs and map them to topic-related queries.

Background read-ahead and task cache. The main agent begins planning and execution while background prediction and retrieval populate the cache:

\begingroup
\def\theHequation{memsidecar-12}
\begin{equation}
\begin{aligned}
\mathcal R_i&=R(\hat x_i,\widehat{\mathcal L}_i,n;\mathcal M),\\
c_i^{\mathrm{cache}}&=(\mathcal R_i,\tau_i^{\mathrm{ready}}).
\end{aligned}
\tag{12}
\end{equation}
\endgroup

The task cache \(C\) stores entry \(c_i^{\mathrm{cache}}\) together with its prediction metadata \((\hat x_i,\widehat{\mathcal L}_i,n)\), where \(\mathcal R_i\) is the actual retrieved entry set and \(\tau_i^{\mathrm{ready}}\) its readiness time. Prediction and read-ahead use an independent execution path. Meanwhile, the main agent can interpret the goal, inspect the environment, or perform preliminary steps as the cache accumulates relevant history.

Actual access and delivery. The main agent forms queries from its task progress. An LLM judges whether the actual and predicted queries express the same memory need; the scheduler then reuses matching, valid cached results.

For a matching prefetched result, let \(\tau_i^{\mathrm{request}}\) be request time and \(\delta_i\) checking and return overhead; \(\tau\) denotes physical time, distinct from discrete step \(t\). Its access wait \(w_i\) is:

\begingroup
\def\theHequation{memsidecar-13}
\begin{equation}
w_i=\max(0,\tau_i^{\mathrm{ready}}-\tau_i^{\mathrm{request}})+\delta_i.
\tag{13}
\end{equation}
\endgroup

A ready result is read directly from cache; an earlier request waits for the remaining preparation. Foresight schedules retrieval ahead of access to support continued execution through time overlap. Requests without a usable match use normal retrieval.

\section{Experiments}\label{experiments}

\begin{table*}[t]
\centering
\small
\renewcommand{\arraystretch}{1.16}
\setlength{\tabcolsep}{3pt}
\renewcommand{\thetable}{1}
\caption{Main results on MemoryArena.}
\label{tab:1}
\begin{tabular*}{\textwidth}{@{\extracolsep{\fill}}l*{8}{c}@{\hspace{10pt}\extracolsep{0pt}}ccc@{}}
\toprule
\raisebox{-.7\baselineskip}{\textbf{Method}} & \multicolumn{2}{c}{\textbf{Shopping}} & \multicolumn{2}{c}{\textbf{Travel}} & \multicolumn{2}{c}{\textbf{\shortstack[c]{Progressive\\Web Search}}} & \multicolumn{2}{c}{\textbf{Reasoning}} & \multicolumn{3}{c}{\textbf{Cost per task $\downarrow$}} \\
\cmidrule(lr){2-3}\cmidrule(lr){4-5}\cmidrule(lr){6-7}\cmidrule(lr){8-9}\cmidrule(l){10-12}
 & SR $\uparrow$ & PS $\uparrow$ & SR $\uparrow$ & PS $\uparrow$ & SR $\uparrow$ & PS $\uparrow$ & SR $\uparrow$ & PS $\uparrow$ & Store (K) & Retrieve (K) & Time (s) \\
\midrule
\multicolumn{12}{@{}l}{\textcolor{MSHeader}{\textit{Memory systems}}} \\
\addlinespace[1pt]
Mem0 & \MSnum{0}{1.33} & \MSnum{0}{48.78} & \MSnum{0}{0.00} & \MSnum{0}{1.02} & \MSnum{0}{11.31} & \MSnum{0}{12.07} & \MSnum{0}{26.67} & \MSnum{0}{47.95} & \MSnum{0}{39.53} & \MSnum{0}{6.43} & \MSnum{0}{789} \\
A-Mem & \MSnum{0}{0.67} & \MSnum{0}{44.78} & \MSnum{0}{0.00} & \MSnum{0}{0.48} & \MSnum{0}{9.95} & \MSnum{0}{9.99} & \MSnum{0}{21.67} & \MSnum{0}{42.95} & \MSnum{0}{55.34} & \MSnum{0}{7.31} & \MSnum{0}{1063} \\
LightMem & \MSnum{0}{1.33} & \MSnum{0}{51.22} & \MSnum{0}{0.00} & \MSnum{0}{1.50} & \MSnum{0}{13.57} & \MSnum{0}{14.99} & \MSnum{0}{28.33} & \MSnum{0}{52.95} & \MSnum{0}{23.72} & \MSnum{0}{7.54} & \MSnum{0}{715} \\
\specialrule{.4pt}{1.8pt}{1.8pt}
\multicolumn{12}{@{}l}{\textcolor{MSHeader}{\textit{Agent harnesses}}} \\
\addlinespace[1pt]
OpenClaw & \MSnum{0}{4.67} & \MSnum{0}{54.67} & \MSnum{0}{1.48} & \MSnum{0}{1.98} & \MSnum{0}{15.84} & \MSnum{0}{17.98} & \MSnum{0}{31.67} & \MSnum{0}{55.91} & \MSnum{1}{15.71} & \MSnum{0}{8.24} & \MSnum{0}{734} \\
ZCode & \MSnum{0}{6.00} & \MSnum{0}{58.11} & \MSnum{0}{2.59} & \MSnum{0}{2.57} & \MSnum{0}{17.65} & \MSnum{0}{20.96} & \MSnum{0}{35.00} & \MSnum{0}{59.77} & \MSnum{0}{18.16} & \MSnum{0}{8.10} & \MSnum{0}{698} \\
DSH & \MSnum{2}{7.33} & \MSnum{0}{59.44} & \MSnum{0}{3.33} & \MSnum{0}{2.51} & \MSnum{0}{18.10} & \MSnum{0}{22.00} & \MSnum{0}{36.67} & \MSnum{0}{60.91} & \MSnum{2}{16.93} & \MSnum{0}{10.39} & \MSnum{0}{686} \\
\specialrule{.4pt}{1.8pt}{1.8pt}
\multicolumn{12}{@{}l}{\textcolor{MSHeader}{\textit{MemSidecar}}} \\
\addlinespace[1pt]
\textbf{Hooks} & \MSnum{2}{7.33} & \MSnum{2}{62.11} & \MSnum{2}{3.70} & \MSnum{2}{2.84} & \MSnum{2}{21.27} & \MSnum{2}{24.01} & \MSnum{2}{43.33} & \MSnum{2}{66.14} & \MSnum{0}{19.61} & \MSnum{2}{6.20} & \MSnum{1}{535} \\
\textbf{Foresight} & \MSnum{1}{8.67} & \MSnum{1}{64.33} & \MSnum{1}{4.81} & \MSnum{1}{3.80} & \MSnum{1}{23.98} & \MSnum{1}{25.41} & \MSnum{1}{48.33} & \MSnum{1}{70.23} & \MSnum{0}{17.85} & \MSnum{1}{6.07} & \MSnum{2}{617} \\
\bottomrule
\end{tabular*}
\par\vspace{3pt}
\begin{minipage}{\textwidth}
\footnotesize\raggedright\textit{Notes.} Each category reports SR / PS (percent; higher is better); reasoning combines mathematics and physics. Store counts memory-maintenance output tokens only; Retrieve counts cumulative memory delivery and prediction tokens. Costs are per complete task, with time in seconds. Hooks and Foresight denote the two independent MemSidecar modes; OpenClaw uses its file-memory subset. Red bold and blue values denote the best and second-best results, respectively; ties share ranks.
\end{minipage}
\end{table*}

\begin{table*}[t]
\centering
\small
\renewcommand{\arraystretch}{1.16}
\setlength{\tabcolsep}{3pt}
\renewcommand{\thetable}{2}
\caption{Main results on MCP-Persona.}
\label{tab:2}
\begin{tabular*}{\textwidth}{@{\extracolsep{\fill}}l*{8}{c}@{\hspace{10pt}\extracolsep{0pt}}ccc@{}}
\toprule
\raisebox{-.7\baselineskip}{\textbf{Method}} & \multicolumn{8}{c}{\textbf{Checkpoint score (\%) $\uparrow$}} & \multicolumn{3}{c}{\textbf{Cost per task $\downarrow$}} \\
\cmidrule(lr){2-9}\cmidrule(l){10-12}
 & CP & CM & SM & EM & Lark & Rednote & Mixed & \textbf{Overall} & Store (K) & Retrieve (K) & Time (s) \\
\midrule
\multicolumn{12}{@{}l}{\textcolor{MSHeader}{\textit{Memory systems}}} \\
\addlinespace[1pt]
Mem0 & \MSnum{0}{32.55} & \MSnum{0}{18.00} & \MSnum{0}{33.14} & \MSnum{0}{43.14} & \MSnum{0}{28.64} & \MSnum{0}{25.54} & \MSnum{0}{6.25} & \MSnum{0}{29.49} & \MSnum{0}{8.02} & \MSnum{0}{3.40} & \MSnum{0}{181.1} \\
A-Mem & \MSnum{0}{27.86} & \MSnum{0}{14.67} & \MSnum{0}{28.49} & \MSnum{0}{39.22} & \MSnum{0}{23.64} & \MSnum{0}{21.20} & \MSnum{0}{6.25} & \MSnum{0}{25.00} & \MSnum{0}{5.32} & \MSnum{0}{4.05} & \MSnum{0}{164.6} \\
LightMem & \MSnum{0}{35.16} & \MSnum{0}{20.67} & \MSnum{0}{35.47} & \MSnum{0}{45.10} & \MSnum{0}{31.59} & \MSnum{0}{27.72} & \MSnum{0}{18.75} & \MSnum{0}{32.18} & \MSnum{0}{3.81} & \MSnum{2}{1.94} & \MSnum{0}{139.8} \\
\specialrule{.4pt}{1.8pt}{1.8pt}
\multicolumn{12}{@{}l}{\textcolor{MSHeader}{\textit{Agent harnesses}}} \\
\addlinespace[1pt]
OpenClaw & \MSnum{0}{37.76} & \MSnum{0}{19.33} & \MSnum{0}{37.79} & \MSnum{0}{47.06} & \MSnum{0}{34.32} & \MSnum{0}{30.43} & \MSnum{0}{18.75} & \MSnum{0}{34.32} & \MSnum{0}{4.91} & \MSnum{0}{3.06} & \MSnum{0}{157.8} \\
ZCode & \MSnum{0}{39.84} & \MSnum{0}{23.33} & \MSnum{0}{39.53} & \MSnum{0}{49.02} & \MSnum{0}{35.23} & \MSnum{0}{28.80} & \MSnum{0}{25.00} & \MSnum{0}{35.77} & \MSnum{0}{6.22} & \MSnum{0}{4.32} & \MSnum{0}{166.9} \\
DSH & \MSnum{0}{40.36} & \MSnum{0}{21.33} & \MSnum{0}{40.12} & \MSnum{0}{46.08} & \MSnum{0}{37.05} & \MSnum{0}{32.61} & \MSnum{0}{18.75} & \MSnum{0}{36.53} & \MSnum{0}{5.33} & \MSnum{0}{3.69} & \MSnum{0}{147.3} \\
\specialrule{.4pt}{1.8pt}{1.8pt}
\multicolumn{12}{@{}l}{\textcolor{MSHeader}{\textit{MemSidecar}}} \\
\addlinespace[1pt]
\textbf{Hooks} & \MSnum{2}{44.79} & \MSnum{2}{26.67} & \MSnum{2}{44.19} & \MSnum{2}{52.94} & \MSnum{2}{41.59} & \MSnum{2}{36.41} & \MSnum{2}{31.25} & \MSnum{2}{41.23} & \MSnum{1}{3.03} & \MSnum{1}{1.72} & \MSnum{1}{82.5} \\
\textbf{Foresight} & \MSnum{1}{48.18} & \MSnum{1}{28.67} & \MSnum{1}{47.09} & \MSnum{1}{55.88} & \MSnum{1}{44.77} & \MSnum{1}{39.67} & \MSnum{1}{37.50} & \MSnum{1}{44.34} & \MSnum{2}{3.11} & \MSnum{0}{2.29} & \MSnum{2}{109.7} \\
\bottomrule
\end{tabular*}
\par\vspace{3pt}
\begin{minipage}{\textwidth}
\footnotesize\raggedright\textit{Notes.} CP, CM, SM, and EM denote collaboration platforms, content management, social media, and email. Lark, Rednote, and Mixed denote cross-server groups. Quality is checkpoint score percentage; Overall is weighted by checkpoint counts. Store and Retrieve are tokens per task, and time is seconds per task. Cost definitions and method abbreviations follow Table 1. Red bold and blue values denote the best and second-best results, respectively; ties share ranks.
\end{minipage}
\end{table*}

\begin{table*}[t]
\centering
\small
\renewcommand{\arraystretch}{1.16}
\setlength{\tabcolsep}{3pt}
\renewcommand{\thetable}{3}
\caption{Dual-layer memory ablations.}
\label{tab:3}
\renewcommand{\arraystretch}{1.24}
\begin{tabular*}{\textwidth}{@{\extracolsep{\fill}}llrrrrrr@{}}
\toprule
\raisebox{-.7\baselineskip}{\textbf{Mode}} & \raisebox{-.7\baselineskip}{\textbf{Configuration}} & \multicolumn{3}{c}{\textcolor{MSHeader}{\textbf{MCP-Persona}}} & \multicolumn{3}{c}{\textcolor{MSHeader}{\textbf{MemoryArena}}} \\
\cmidrule(lr){3-5}\cmidrule(l){6-8}
 & & Overall (\%) $\uparrow$ & Store (K) $\downarrow$ & Retrieve (K) $\downarrow$ & SR (\%) $\uparrow$ & Store (K) $\downarrow$ & Retrieve (K) $\downarrow$ \\
\midrule
\textbf{Hooks} & \textbf{Full} & \textbf{41.23} & \textbf{3.03} & \textbf{1.72} & \textbf{13.41} & \textbf{19.61} & \textbf{6.20} \\
 & w/o W & 36.46\, {\scriptsize\textcolor{MSDeltaBad}{(-4.77)}} & 3.30\, {\scriptsize\textcolor{MSDeltaBad}{(+0.27)}} & 1.92\, {\scriptsize\textcolor{MSDeltaBad}{(+0.20)}} & 10.70\, {\scriptsize\textcolor{MSDeltaBad}{(-2.71)}} & 21.29\, {\scriptsize\textcolor{MSDeltaBad}{(+1.68)}} & 6.78\, {\scriptsize\textcolor{MSDeltaBad}{(+0.58)}} \\
 & w/o G & 38.67\, {\scriptsize\textcolor{MSDeltaBad}{(-2.56)}} & 3.21\, {\scriptsize\textcolor{MSDeltaBad}{(+0.18)}} & 1.79\, {\scriptsize\textcolor{MSDeltaBad}{(+0.07)}} & 9.84\, {\scriptsize\textcolor{MSDeltaBad}{(-3.57)}} & 20.81\, {\scriptsize\textcolor{MSDeltaBad}{(+1.19)}} & 6.81\, {\scriptsize\textcolor{MSDeltaBad}{(+0.61)}} \\
\specialrule{.4pt}{1.8pt}{1.8pt}
\textbf{Foresight} & \textbf{Full} & \textbf{44.34} & \textbf{3.11} & \textbf{2.29} & \textbf{15.41} & \textbf{17.85} & \textbf{6.07} \\
 & w/o W & 38.12\, {\scriptsize\textcolor{MSDeltaBad}{(-6.22)}} & 3.37\, {\scriptsize\textcolor{MSDeltaBad}{(+0.25)}} & 2.59\, {\scriptsize\textcolor{MSDeltaBad}{(+0.31)}} & 10.98\, {\scriptsize\textcolor{MSDeltaBad}{(-4.42)}} & 19.32\, {\scriptsize\textcolor{MSDeltaBad}{(+1.47)}} & 6.60\, {\scriptsize\textcolor{MSDeltaBad}{(+0.53)}} \\
 & w/o G & 39.92\, {\scriptsize\textcolor{MSDeltaBad}{(-4.42)}} & 3.29\, {\scriptsize\textcolor{MSDeltaBad}{(+0.18)}} & 2.41\, {\scriptsize\textcolor{MSDeltaBad}{(+0.12)}} & 10.27\, {\scriptsize\textcolor{MSDeltaBad}{(-5.14)}} & 19.38\, {\scriptsize\textcolor{MSDeltaBad}{(+1.53)}} & 6.62\, {\scriptsize\textcolor{MSDeltaBad}{(+0.56)}} \\
\bottomrule
\end{tabular*}
\par\vspace{3pt}
\begin{minipage}{\textwidth}
\footnotesize\raggedright\textit{Notes.} Full retains both layers; w/o W removes Workspace and w/o G removes Global. Metrics follow the main tables: MCP-Persona Overall, MemoryArena task SR, and Store/Retrieve tokens. Parentheses show changes from the same-mode Full: quality in percentage points and tokens in K (1,000 tokens).
\end{minipage}
\end{table*}

\begin{figure*}[t]
\centering
\includegraphics[width=\textwidth]{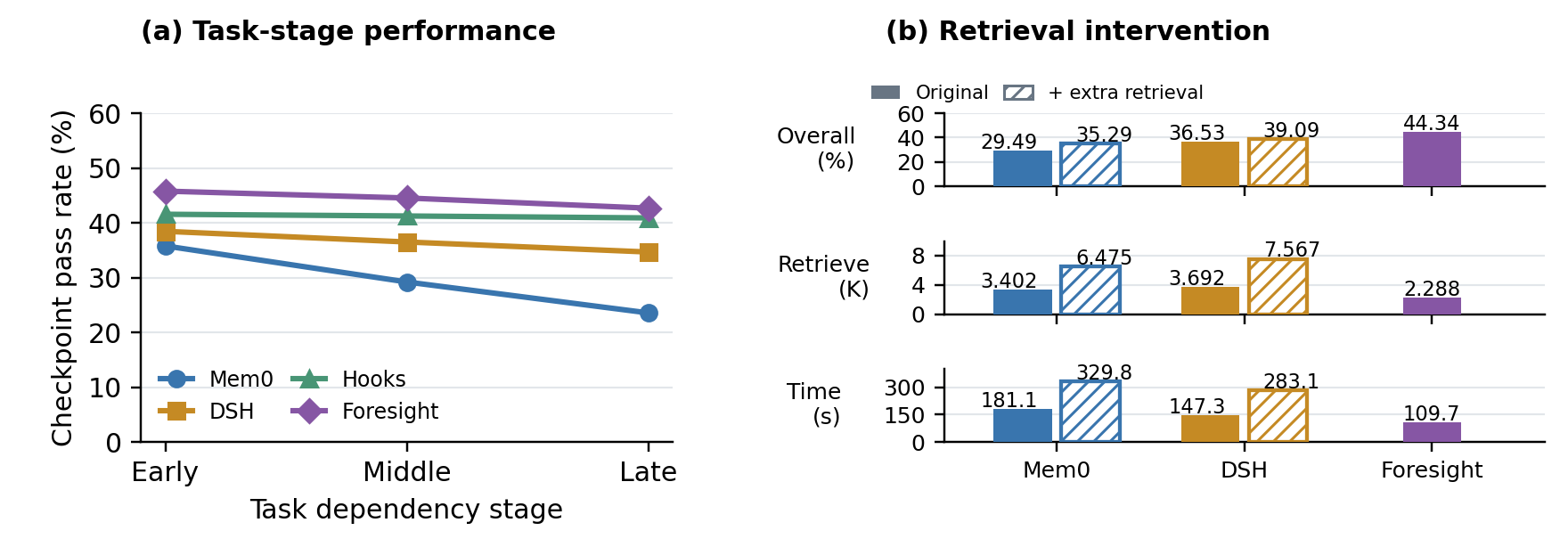}
\caption{MCP-Persona stage performance and retrieval intervention. (a) Early-, middle-, and late-stage checkpoint pass rates. (b) Original and extra-retrieval configurations of Mem0 and DSH, with standard Foresight as a reference; bars report Overall (\%), Retrieve tokens (thousands), and runtime (seconds).}
\label{fig:3}
\end{figure*}

We compare MemSidecar with traditional memory systems and harness memory policies on MemoryArena and MCP-Persona, evaluating task quality and execution efficiency. Memory-layer ablations, task-stage analysis, retrieval intervention, and a case study examine the roles of memory organization and scheduling. Hooks and Foresight are evaluated independently.

\subsection{Experimental Setup}\label{experimental-setup}

\subsubsection{Evaluation Tasks}\label{evaluation-tasks}

MemoryArena organizes dependent tasks across sessions \cite{he2026memoryarena}. It contains 701 task sequences and 4,850 stages, covering bundled web shopping, group travel planning, Progressive Web Search, and mathematical and physical reasoning. Later stages require earlier shopping requirements, group arrangements, search evidence, or reasoning progress.

MCP-Persona provides stateful simulators for personal applications \cite{wang2026mcppersona}. Its 173 English tasks contain 724 checkpoints and cover collaboration platforms (CP), content management (CM), social media (SM), email (EM), and cross-server Lark, Rednote, and Mixed tasks. Agents must use historical information to perform queries, content management, and collaboration across applications.

\subsubsection{Compared Methods}\label{compared-methods}

As agent harnesses connect persistent memory to session management, context assembly, and tool calls, runtime frameworks have become another way to provide memory capabilities. We therefore compare traditional memory systems and harness memory policies, covering both independent memory components and mechanisms integrated into host execution flows.

Traditional memory systems provide memory organization, writing, and retrieval as separate components.

\begin{itemize}
\tightlist
\item
  Mem0: fact extraction, consolidation, and vector retrieval \cite{chhikara2025mem0}.
\item
  A-Mem: structured notes, associative links, and memory evolution \cite{xu2025amem}.
\item
  LightMem: sensory, short-term, and long-term stages that compress and organize information, with offline long-term consolidation \cite{fang2026lightmem}.
\end{itemize}

Harness memory policies connect persistence to host sessions, context management, or tool calls.

\begin{itemize}
\tightlist
\item
  OpenClaw: file memory and dated logs associated with sessions; the current comparison uses its file-memory subset \cite{openclaw2026memory}.
\item
  ZCode: project-level Memory that consolidates useful information after conversations and loads the index in later sessions within the same project \cite{zai2026zcode}.
\item
  DSH: the DeepSeek Harness MCP-memory integration; this protocol selects the official MCP Reference Memory example, which exposes entity, relation, and observation storage together with graph-memory access and search \cite{deepseek2026memory}.
\end{itemize}

These methods are compared with MemSidecar-\allowbreak{}Hooks and MemSidecar-\allowbreak{}Foresight using a common task executor.

\subsubsection{Models and Execution Configuration}\label{models-and-execution-configuration}

GPT-5.4-mini executes the main tasks, while Qwen3-8B handles memory management, retrieval, and query prediction. Methods use the same task inputs, environment tools, and execution budgets within each dataset.

Hooks delivers memories under execution events and conditions, while Foresight uses an independent predictor to prepare a task-level cache. Appendix B.1 gives retrieval and scheduling settings. Memory-layer ablations retain the remaining configuration and budgets of each mode.

\subsubsection{Metrics}\label{metrics}

Task quality. MemoryArena reports complete-task success rate (SR) and stage score rate (PS) for each of four categories. SR aggregates fully successful tasks, and PS aggregates stage scores. The reasoning category combines mathematics and physics.

MCP-Persona reports checkpoint score rates for each category. Overall aggregates checkpoint scores across all categories.

Execution efficiency. Store Token counts model outputs for memory extraction, writing, updating, and Hook predicate generation. Retrieve Token includes cumulative memory text and schema inputs delivered to the main model plus prediction inputs and outputs. Runtime is end-to-end seconds per complete task. Costs are reported per task and weighted by task counts across categories.

\subsection{Main Experiments}\label{main-experiments}

\subsubsection{MemoryArena}\label{memoryarena}

Table 1 reports quality separately for shopping, travel, search, and reasoning. Foresight obtains SR values of 8.67\%, 23.98\%, and 48.33\% on shopping, search, and reasoning, with PS values of 64.33\%, 25.41\%, and 70.23\%. Hooks obtains SR values of 7.33\%, 21.27\%, and 43.33\% and PS values of 62.11\%, 24.01\%, and 66.14\%. Both modes exceed the strongest external baseline on PS and at least match it on SR in all four categories.

Complete-task success remains low in travel planning. Hooks and Foresight obtain SR values of 3.70\% and 4.81\%, with PS values of 2.84\% and 3.80\%, respectively. Foresight exceeds Hooks on both complete-task success rate and stage score rate.

Hooks uses 19,612 Store tokens and 6,201 Retrieve tokens with a time of 535 seconds; Foresight uses 17,852 and 6,069 tokens with a time of 617 seconds. Foresight uses fewer tokens in both memory-cost columns, while Hooks finishes sooner.

\subsubsection{MCP-Persona}\label{mcp-persona}

In Table 2, Hooks and Foresight obtain Overall scores of 41.23\% and \textbf{44.34\%}. The strongest external baseline, DSH, scores 36.53\%; Hooks exceeds it by 4.70 percentage points, and Foresight exceeds it by \textbf{7.80 percentage points}. Foresight leads 7 task groups. On Mixed, Hooks and Foresight score 31.25\% and 37.50\%, compared with 25.00\% for ZCode, the strongest external baseline in this group.

Hooks uses 3,032 Store tokens and 1,720 Retrieve tokens with a time of \textbf{82.5 seconds}; Foresight uses 3,114 and 2,288 tokens with a time of 109.7 seconds. Hooks has the lowest value in all three cost columns. Foresight is also below DSH, the strongest external baseline on quality, in all three cost columns.

\subsection{Memory Ablations}\label{memory-ablations}

We remove Workspace (w/o W) or Global (w/o G) separately in Hooks and Foresight to examine the roles of both memory layers. Removing a layer disables its writes and reads. Other models, scheduling policies, and budgets remain fixed.

Table 3 shows different dependencies on the two memory layers. On MCP-Persona, retaining only Workspace yields Overall scores of 38.67\% for Hooks and 39.92\% for Foresight, both above their Global-only counterparts. MemoryArena shows the opposite pattern: Global-only Hooks and Foresight reach 10.70\% and 10.98\%, exceeding the corresponding Workspace-only results of 9.84\% and 10.27\%. Removing either layer increases Store and Retrieve tokens relative to Full in both modes, indicating that the full dual-layer memory configuration benefits both task performance and memory efficiency.

\subsection{Long-Horizon Performance and Retrieval Intervention}\label{long-horizon-performance-and-retrieval-intervention}

We examine memory scheduling on MCP-Persona through task-stage performance and additional retrieval, comparing Mem0, DSH, Hooks, and Foresight.

\subsubsection{Performance across task stages}\label{performance-across-task-stages}

Figure 3(a) groups checkpoints into early, middle, and late stages according to their execution-step positions and reports the pass rate for each stage. Mem0 declines from 35.8\% to 23.6\%, a drop of 12.2 percentage points; DSH, Hooks, and Foresight decline by 3.8, 0.7, and 3.1 points, respectively. As execution progresses, earlier constraints, user preferences, and tool history leave active context, making later decisions more dependent on memory recall. Mem0's marked decline reflects insufficient use of historical information in later stages. Hooks has the flattest curve, while Foresight achieves the highest pass rate at every stage, showing how continued delivery of relevant memories supports long-horizon execution.

\subsubsection{Additional retrieval at specified events}\label{additional-retrieval-at-specified-events}

For actions requiring personalized information, additional retrieval is performed at the corresponding memory-delivery points. Figure 3(b) compares the extra-retrieval and original configurations of Mem0 and DSH, with standard Foresight as a reference. The three bar subpanels report Overall checkpoint scores, Retrieve tokens, and runtime.

Additional retrieval increases Mem0's Overall from 29.49\% to 35.29\% and DSH's from 36.53\% to 39.09\%, gains of \textbf{5.80} and \textbf{2.56 percentage points}. Retrieve tokens rise from 3,402 to 6,475 for Mem0 and from 3,692 to 7,567 for DSH, increases of 90.3\% and 105.0\%, respectively. Runtime rises from 181.1 to 329.8 seconds for Mem0 and from 147.3 to 283.1 seconds for DSH, increases of 82.1\% and 92.2\%, respectively. Foresight achieves 44.34\% Overall with 2,288 Retrieve tokens and 109.7 seconds. Extra queries improve baseline scores while substantially increasing retrieval usage and execution time.

\subsection{Case Study}\label{case-study}

Figure 4 presents the execution trace of MCP-Persona task 153: find a route to a client's office, post a Slack meeting notice, check Feishu participants, and react to the new notice. Global Memory supplies existing addresses and group information; Workspace Memory retains the new channel, message timestamp, and pending reaction.

\begin{figure}[htbp]
\centering
\includegraphics[width=.90\columnwidth]{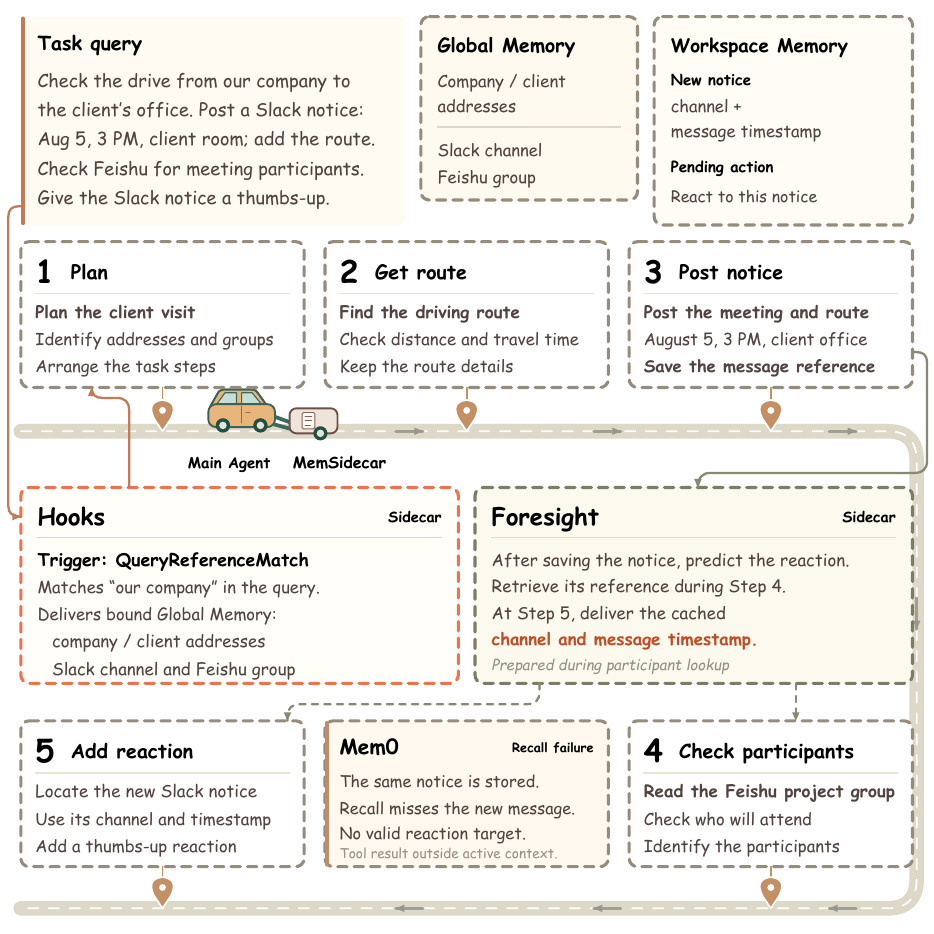}
\caption{Execution trace of memory scheduling across stages in MCP-Persona task 153.}
\label{fig:4}
\end{figure}

Hooks binds reference conditions such as ``our company'' to the associated memories. When the initial request satisfies them, the host directly delivers company/client addresses and project-group information for route planning and posting. After the Slack notice is posted, its state is written to Workspace. Foresight prefetches this state while the main agent checks Feishu participants, then delivers the message reference when the actual query arrives at Step 5, enabling the reaction. Mem0 stores the notice state but fails to recall its reference after the earlier tool result leaves active context, leaving the subsequent action without its target message.

\section{Conclusion}\label{conclusion}

We formulate memory scheduling as identifying needs and coordinating memory preparation and delivery during long-horizon execution. MemSidecar separates these decisions from task execution through two independently selectable modes. Hooks delivers bound memory when conditions match. Foresight predicts upcoming queries and prepares results in the background. Experiments on MemoryArena and MCP-Persona show improved accuracy and efficiency, with shared-backend comparisons further supporting the value of scheduling. Memory scheduling is therefore an important issue in designing long-horizon agents, and MemSidecar offers a viable approach to improving long-horizon task execution through independent memory scheduling.

\FloatBarrier
\label{end:main}
\clearpage
\twocolumn
\setlength{\bibsep}{-0.5pt}
\bibliographystyle{ACM-Reference-Format}
\bibliography{generated/references}
\clearpage
\appendix
\section{Framework Details}\label{framework-details}

\subsection{Memory Representation and Updating}\label{memory-representation-and-updating}

At a stage or task write boundary, the maintenance input combines the sets of observed interactions \(H_t^{\mathrm{obs}}\) and tool results \(\mathcal O_t^{\mathrm{obs}}\). Extraction organizes reusable facts or coherent task state into candidate entries:

\begingroup
\def\theHequation{memsidecar-A1}
\begin{equation}
\mathcal D_t=H_t^{\mathrm{obs}}\cup\mathcal O_t^{\mathrm{obs}},\qquad \mathcal P_t=\operatorname{Extract}(\mathcal D_t).
\tag{A1}
\end{equation}
\endgroup

Each candidate contains content \(c\), memory layer \(\ell\), and applicability \(\omega\). Content can describe a confirmed constraint, environmental fact, completed step, or pending dependency. Applicability identifies the relevant object, project, and time range:

\begingroup
\def\theHequation{memsidecar-A2}
\begin{equation}
m=(c,\ell,\omega),\qquad \ell\in\{\mathrm{G},\mathrm{W}\}.
\tag{A2}
\end{equation}
\endgroup

General knowledge, user preferences, and cross-task conventions enter Global Memory. Project procedures, workspace conditions, and progress enter Workspace Memory. One-off requirements remain with the current task, while user preferences carry into later tasks that meet their applicability conditions. Global Memory connects content through topic and knowledge nodes; Workspace Memory links operational experience with current state.

Deduplication compares topic, factual object, and applicability within the assigned layer to locate an existing counterpart:

\begingroup
\def\theHequation{memsidecar-A3}
\begin{equation}
j^*(m)=\operatorname{Match}(c,\omega,\mathcal M_{\ell,t}).
\tag{A3}
\end{equation}
\endgroup

\(\operatorname{Match}\) returns the identifier of an existing counterpart, or \(\bot\) when no match exists. An unmatched candidate is inserted. Matching content retains its wording; new observations fill missing facts or replace changed state. Conflicts are resolved within their applicability scope: current explicit requirements override conflicting historical preferences, while entries valid under different conditions remain separate. Extraction, deduplication, and necessary updates form one maintenance step:

\begingroup
\def\theHequation{memsidecar-A4}
\begin{equation}
\mathcal M_{\ell,t+1}=\operatorname{Upsert}(\mathcal M_{\ell,t},\mathcal P_{\ell,t}),\qquad \ell\in\{\mathrm{G},\mathrm{W}\}.
\tag{A4}
\end{equation}
\endgroup

Here \(\mathcal P_{\ell,t}\) contains candidates assigned to layer \(\ell\), and \(t\) retains the execution-step index used in the main text. Equation (A4) updates memory only at stage or task write boundaries; at other steps, \(\mathcal M_{\ell,t+1}=\mathcal M_{\ell,t}\).

\subsection{Complete Scheduling Algorithms}\label{complete-scheduling-algorithms}

Algorithm A1 checks delivery conditions on a new query or before tool execution; successfully delivered Hooks do not fire again within that task. In the algorithm, fired contains successfully delivered Hook indices, and used contains the Hook indices corresponding to the packed memory content; used joins fired only after successful delivery, and both entry points share the latest set. Algorithm A2 predicts queries and prepares a cache in the background. An LLM matches actual queries to predicted memory needs, reusing valid results or retrieving normally otherwise. Observed writes refresh the visible directory, and background preparation uses the remaining task-wide budget. Both modes maintain observed memory at stage/task boundaries.

\begin{figure*}[t]
\centering
\begin{minipage}[t]{.485\textwidth}
\vspace{0pt}
\begin{algorithm}[H]
\small
\raggedright
\renewcommand{\thealgorithm}{A1}
\captionsetup{hypcap=false,type=algorithm}
\caption{Hooks-based task execution.}
\label{alg:A1}
\begin{algorithmic}[1]
\Require instruction q; memory M; memory-bound Hooks hooks
\Ensure task result y; updated memory M
\Function{DeliverHooks}{event, state, action}
\State matches \(\gets\) UnfiredMatches(hooks, fired, event, state, action)
\State (text, used) \(\gets\) PackBoundEntries(matches, M)
\State (state, sent) \(\gets\) AppendContext(state, text)
\If{sent}
\State fired \(\gets\) fired union used
\State \Return state, text
\EndIf
\State \Return state, empty
\EndFunction
\State (task, state) \(\gets\) BeginTask(q)
\State fired \(\gets\) empty
\While{task is active and execution budget remains}
\State (state, \_) \(\gets\) DeliverHooks(QueryEvent(state), state, none)
\State action \(\gets\) MainAgent(state)
\State (state, context) \(\gets\) DeliverHooks(PreTool, state, action)
\If{context is not empty}
\State action \(\gets\) MainAgent.Revise(state, action, context)
\EndIf
\State observation \(\gets\) ExecuteTaskStepOnce(action)
\State state \(\gets\) Advance(state, action, observation)
\If{a stage or task write boundary is reached}
\State (M, hooks) \(\gets\) MaintainMemoriesAndHooks(M, hooks, state)
\EndIf
\EndWhile
\State CloseTask(task)
\State \Return Result(state), M
\end{algorithmic}
\par\vspace{3pt}
\begin{minipage}{\columnwidth}
\footnotesize Query and PreTool share task-local trigger marks; memory content is merged and deduplicated, then delivered within the common budget.
\end{minipage}
\end{algorithm}
\end{minipage}
\hfill
\begin{minipage}[t]{.485\textwidth}
\vspace{0pt}
\begin{algorithm}[H]
\small
\raggedright
\renewcommand{\thealgorithm}{A2}
\captionsetup{hypcap=false,type=algorithm}
\caption{Foresight-based task execution.}
\label{alg:A2}
\begin{algorithmic}[1]
\Require instruction q; memory M; retriever R; predictor F; budget K
\Ensure task result; updated memory M
\Function{Refill}{}
\State queries \(\gets\) PredictNew(F, q, topics, Kremaining)
\State Kremaining \(\gets\) Kremaining - Count(queries)
\State RetrieveAndCache(R, queries, C)
\EndFunction
\State (task, state, C) \(\gets\) BeginTask(q); Kremaining \(\gets\) K
\State topics \(\gets\) VisibleDirectory(M)
\State StartBackground(Refill)
\While{task is active and execution budget remains}
\State action \(\gets\) MainAgent(state)
\If{action is a memory query (x, layers, n)}
\State memories \(\gets\) LLMMatchOrWait(C, x, layers, n)
\If{memories are unavailable}
\State memories \(\gets\) R(x, layers, n; M)
\EndIf
\State state \(\gets\) Deliver(state, PackUsefulText(memories))
\Else
\State state \(\gets\) Advance(state, action, Execute(action))
\If{a stage or task write boundary is reached}
\State M \(\gets\) MaintainObserved(M, state, C)
\State topics \(\gets\) VisibleDirectory(M); WakeBackground(Refill)
\EndIf
\EndIf
\EndWhile
\State CloseTaskAndCache(task, C)
\State \Return Result(state), M
\end{algorithmic}
\par\vspace{3pt}
\begin{minipage}{\columnwidth}
\footnotesize Background preparation shares the task-wide budget; actual queries use matching cache results or normal retrieval.
\end{minipage}
\end{algorithm}
\end{minipage}
\end{figure*}

\section{Experimental Details and Additional Experiments}\label{experimental-details-and-additional-experiments}

\subsection{Experimental Details}\label{experimental-details}

GPT-5.4-mini executes the main tasks, and Qwen3-8B handles memory management, retrieval, and prediction. The two benchmarks are executed and reported separately. Methods share visible inputs, task tools, and execution budgets, build independent memory stores, and follow the same task order. MemSidecar encodes queries and memory text with OpenAI \nolinkurl{text-embedding-3-small}, retrieves at most six entries per query, and delivers at most 6,000 characters. Foresight uses a task-wide prediction budget of \(K=4\), a cache lifetime of 600 seconds, and a task lifetime of 3,600 seconds. Hooks uses predicates generated by the memory model; both modes share the delivery budget.

The current configuration primarily matches user-query content and context, or the name and arguments of a pending action. Each Hook delivers at most once per task: \(\mathcal F_t\) contains indices of successfully delivered Hooks and updates immediately after each successful delivery. Both entry points share its latest state to prevent redelivery. The set resets for a new task. Memory updates maintain Hook bindings.

Metrics follow Section 4.1: MemoryArena reports SR/PS for each task category, and MCP-Persona reports category-level checkpoint scores and Overall. Store counts outputs for memory maintenance and Hook predicate generation, Retrieve counts cumulative memory-delivery and prediction tokens, and Runtime measures end-to-end time for the full online workflow, all normalized per complete task. Memory ablations report MCP-Persona Overall, MemoryArena complete-task SR, and Store/Retrieve tokens; additional filtering costs enter Retrieve.

\subsection{Training ForesightAgent}\label{training-foresightagent}

This experiment examines whether ForesightAgent can be improved through training. We compare untrained, SFT, and SFT+RL configurations to assess whether supervision and task-level rewards improve downstream task performance, alongside the associated Retrieve Token and Runtime costs. Training examples are constructed from MemoryArena task questions, using GPT-5.6-sol canonical answers as supervision. After training, we freeze ForesightAgent and evaluate it on MCP-Persona.

The untrained condition reuses the main-table Foresight configuration. SFT trains for 50 steps, learning demand queries from task instructions and visible topics. SFT+RL initializes from the SFT parameters and continues reinforcement learning for 50 steps, with \(+1\) for a correct or successful complete task and \(0\) otherwise. Training uses a batch size of 32. Only ForesightAgent is trained; the main agent, memory maintenance, retriever, and cache rules remain fixed.

\begin{table}[!htbp]
\centering
\small
\renewcommand{\arraystretch}{1.16}
\setlength{\tabcolsep}{3pt}
\renewcommand{\thetable}{B1}
\caption{ForesightAgent training comparison on MCP-Persona.}
\label{tab:B1}
\captionsetup{justification=raggedright,singlelinecheck=false}
\renewcommand{\arraystretch}{1.22}
\setlength{\tabcolsep}{4pt}
\begin{tabularx}{\columnwidth}{@{}>{\hsize=1.05\hsize\linewidth=\hsize\raggedright\arraybackslash}X>{\hsize=1.05\hsize\linewidth=\hsize\centering\arraybackslash}X>{\hsize=1.1\hsize\linewidth=\hsize\centering\arraybackslash}X>{\hsize=0.8\hsize\linewidth=\hsize\centering\arraybackslash}X@{}}
\toprule
\textbf{Condition} & \textbf{Overall (\%)}~$\uparrow$ & \textbf{Retrieve (K)}~$\downarrow$ & \textbf{Time (s)}~$\downarrow$ \\
\midrule
Untrained & 44.34 & 2.29 & 109.7 \\
SFT & 46.27 & 2.58 & 119.8 \\
SFT+RL & 49.17 & 2.82 & 113.7 \\
\bottomrule
\end{tabularx}
\par\vspace{3pt}
\begin{minipage}{\columnwidth}
\footnotesize\raggedright\textit{Notes.} Overall is the checkpoint score percentage. Retrieve Token and Runtime follow the main-table definitions; time is in seconds.
\end{minipage}
\end{table}

In Table B1, SFT raises Overall from 44.34\% to 46.27\%, an increase of 1.93 percentage points over the untrained condition. Further reinforcement learning reaches 49.17\%, adding 2.90 points over SFT and 4.83 points over the untrained predictor. Retrieve Token is 2.58K for SFT and 2.82K for SFT+RL, increases of 0.30K and 0.53K over the baseline. Runtime increases by 10.1 and 4.0 seconds, respectively. These results support the feasibility of improving ForesightAgent through training, with higher downstream scores accompanied by additional online costs.

\subsection{Memory Filtering and Joint Scheduling}\label{memory-filtering-and-joint-scheduling}

This experiment asks whether Hooks and Foresight can be combined as complementary scheduling methods within the same task. Condition-based delivery and predicted queries cover memory needs in different ways. We compare joint scheduling with each standalone mode to assess whether combining them improves task quality and what additional Retrieve Token and Runtime it requires. Hooks+filtering provides an additional comparison that checks the applicability of Hook-matched entries without enabling Foresight.

We compare only Hooks, only Foresight, Hooks+filtering, and Hooks+Foresight joint delivery. Filtering checks the applicability of content selected for delivery. Joint delivery enables event-based delivery and predicted queries, merging and deduplicating the two streams within the shared budget.

\begin{table*}[t]
\centering
\small
\renewcommand{\arraystretch}{1.16}
\setlength{\tabcolsep}{3pt}
\renewcommand{\thetable}{B2}
\caption{Single-mode, filtering, and joint-delivery comparison.}
\label{tab:B2}
\renewcommand{\arraystretch}{1.18}
\setlength{\tabcolsep}{3pt}
\begin{tabular*}{\textwidth}{@{\extracolsep{\fill}}l*{4}{c}@{\hspace{12pt}}*{4}{c}@{}}
\toprule
\raisebox{-.65\baselineskip}{\textbf{Condition}} & \multicolumn{4}{c}{\textcolor{MSHeader}{\textbf{MCP-Persona}}} & \multicolumn{4}{c}{\textcolor{MSHeader}{\textbf{MemoryArena}}} \\
\cmidrule(lr){2-5}\cmidrule(l){6-9}
 & Overall (\%)~$\uparrow$ & Store (K)~$\downarrow$ & Retrieve (K)~$\downarrow$ & Time (s)~$\downarrow$ & SR (\%)~$\uparrow$ & Store (K)~$\downarrow$ & Retrieve (K)~$\downarrow$ & Time (s)~$\downarrow$ \\
\midrule
only Hooks & 41.23 & \textbf{3.03} & 1.72 & \textbf{82.5} & 13.41 & 19.61 & 6.20 & \textbf{535} \\
only Foresight & 44.34 & 3.11 & 2.29 & 109.7 & 15.41 & \textbf{17.85} & 6.07 & 617 \\
Hooks+filtering & 42.13 & \textbf{3.03} & \textbf{1.65} & 91.6 & 14.12 & 19.61 & \textbf{5.74} & 585 \\
Hooks+Foresight & \textbf{46.13} & 3.18 & 2.48 & 122.3 & \textbf{16.12} & 21.14 & 7.29 & 653 \\
\bottomrule
\end{tabular*}
\par\vspace{3pt}
\begin{minipage}{\textwidth}
\footnotesize\raggedright\textit{Notes.} Single-mode results follow Tables 1–2. MCP-Persona reports the Overall checkpoint score, and MemoryArena reports complete-task SR. Store, Retrieve, and time are per complete task.
\end{minipage}
\end{table*}

\begin{table*}[t]
\centering
\small
\renewcommand{\arraystretch}{1.16}
\setlength{\tabcolsep}{3pt}
\renewcommand{\thetable}{B3}
\caption{Scheduling with a shared backend.}
\label{tab:B3}
\renewcommand{\arraystretch}{1.18}
\setlength{\tabcolsep}{3pt}
\begin{tabular*}{\textwidth}{@{\extracolsep{\fill}}l*{3}{c}@{\hspace{12pt}}*{3}{c}@{}}
\toprule
\raisebox{-.65\baselineskip}{\textbf{Method}} & \multicolumn{3}{c}{\textcolor{MSHeader}{\textbf{MCP-Persona}}} & \multicolumn{3}{c}{\textcolor{MSHeader}{\textbf{MemoryArena}}} \\
\cmidrule(lr){2-4}\cmidrule(l){5-7}
 & Overall (\%)~$\uparrow$ & Retrieve (K)~$\downarrow$ & Time (s)~$\downarrow$ & SR (\%)~$\uparrow$ & Retrieve (K)~$\downarrow$ & Time (s)~$\downarrow$ \\
\midrule
Mem0 & 35.70 & 2.65 & 148.6 & 10.27 & 6.87 & 747 \\
DSH & 39.16 & 2.91 & 136.8 & 12.55 & 8.13 & 703 \\
Hooks & 41.23 & \textbf{1.72} & \textbf{82.5} & 13.41 & 6.20 & \textbf{535} \\
Foresight & \textbf{44.34} & 2.29 & 109.7 & \textbf{15.41} & \textbf{6.07} & 617 \\
\bottomrule
\end{tabular*}
\par\vspace{3pt}
\begin{minipage}{\textwidth}
\footnotesize\raggedright\textit{Notes.} MCP-Persona reports Overall; MemoryArena reports SR. Tokens and time are per task.
\end{minipage}
\end{table*}

On MCP-Persona, Hooks+filtering gains 0.90 percentage points over only Hooks, reduces Retrieve Token by 0.07K, and adds 9.1 seconds of Runtime. Filtering trades additional time for more selective delivery and a higher score. Joint scheduling reaches 46.13\% Overall, exceeding only Foresight by 1.80 points while adding 0.19K Retrieve Token and 12.6 seconds. Joint scheduling improves task scores and increases preparation and delivery costs.

On MemoryArena, Hooks+filtering gains 0.71 percentage points over only Hooks, reduces Retrieve Token by 0.46K, and adds 50 seconds of Runtime. Joint scheduling reaches 16.12\% SR, exceeding only Foresight by 0.71 points while adding 3.29K Store tokens, 1.22K Retrieve Token, and 36 seconds. Joint delivery increases success rate and incurs additional memory maintenance, preparation, and delivery costs.

\subsection{Scheduling with a Shared Backend}\label{scheduling-with-a-shared-backend}

With the memory backend fixed, we examine the efficiency of the scheduling policies through Retrieve Token and Runtime, alongside task scores. The four methods share the same Global Memory/Workspace Memory backend and experimental settings, retaining only their respective access and scheduling policies. Hooks and untrained Foresight reuse the main-table configurations and results.

With the memory backend held fixed, both MemSidecar modes use fewer Retrieve tokens and less time than Mem0 and DSH while achieving higher task scores, supporting the efficiency benefits of the scheduling policies themselves. Hooks has the shortest runtime and Foresight the highest quality on both datasets. Relative to DSH, Foresight gains 5.18 percentage points on MCP-Persona and 2.85 points on MemoryArena.

\end{document}